\documentclass[10pt, a4paper]{article}

\usepackage[final]{lrec2026} %
\usepackage{times}
\usepackage{latexsym}

\usepackage[T1]{fontenc}

\usepackage[utf8]{inputenc}

\usepackage{microtype}

\usepackage{inconsolata}

\usepackage{graphicx}

\usepackage{todonotes}

\usepackage{float}

\usepackage{subcaption,booktabs,multirow,comment,graphicx} %
\usepackage{tabularx}
\usepackage[capitalise]{cleveref}
\usepackage{xcolor}
\usepackage{tcolorbox}
\usepackage{booktabs}

\title{Automatic Essay Scoring and Feedback Generation in Basque Language Learning}

\name{Ekhi Azurmendi, Xabier Arregi, Oier Lopez de Lacalle} 

\address{HiTZ Center - Ixa, University of the Basque Country UPV/EHU \\
         \{ekhi.azurmendi\}@ehu.eus\\}

\abstract{
This paper introduces the first publicly available dataset for Automatic Essay Scoring (AES) and feedback generation in Basque, targeting the CEFR C1 proficiency level. The dataset comprises 3,200 essays from HABE, each annotated by expert evaluators with criterion specific scores covering correctness, richness, coherence, cohesion, and task alignment enriched with detailed feedback and error examples. We fine-tune open-source models, including RoBERTa-EusCrawl and Latxa 8B/70B, for scoring. We focused on correctness criteria for the explanation generation, adapting Latxa to correctly predict both, scores and explanations. Our experiments show that encoder models remain highly reliable for AES, while supervised fine-tuning (SFT) of Latxa significantly enhances performance, surpassing state-of-the-art (SoTA) closed-source systems such as GPT-5 and Claude Sonnet 4.5 in scoring consistency and feedback quality. We also propose a novel evaluation methodology for assessing feedback generation, combining automatic consistency metrics with expert-based validation of extracted learner errors. Results demonstrate that the fine-tuned Latxa model produces criterion-aligned, pedagogically meaningful feedback and identifies a wider range of error types than proprietary models. This resource and benchmark establish a foundation for transparent, reproducible, and educationally grounded NLP research in low-resource languages such as Basque. The dataset, models, and manual evaluation annotations are available at \url{https://huggingface.co/collections/EkhiAzur/habe-hitz-c1}.
 \\ \newline  }

\begin{document}

\maketitleabstract

\section{Introduction}

The assessment of advanced second-language writing remains a complex and resource-intensive task, particularly at CEFR Level C1, where proficiency is characterized by near-native use of language, flexibility in style and register, and coherence across extended discourse. While traditional human-based evaluation offers rich pedagogical insights, it is inherently time-consuming, costly, and susceptible to inter-rater variability, thus limiting its scalability in large-scale or continuous learning contexts~\cite{arriola-etal-2023-towards}. \textbf{Automatic Essay Scoring} (AES) offers a promising alternative by enabling consistent and scalable evaluation~\cite{ke2019automated}. Recent advances in natural language processing (NLP) and large language models (LLM) have substantially improved the reliability of AES systems. However, scoring alone provides limited pedagogical benefit if learners are not informed of the specific linguistic aspects underlying their performance. Most existing AES approaches perform a holistic scoring, summarizing the quality of an essay with a single score~\cite{wang2023aggregating}.

\begin{figure}[t]
    \centering
    \includegraphics[width=1\linewidth]{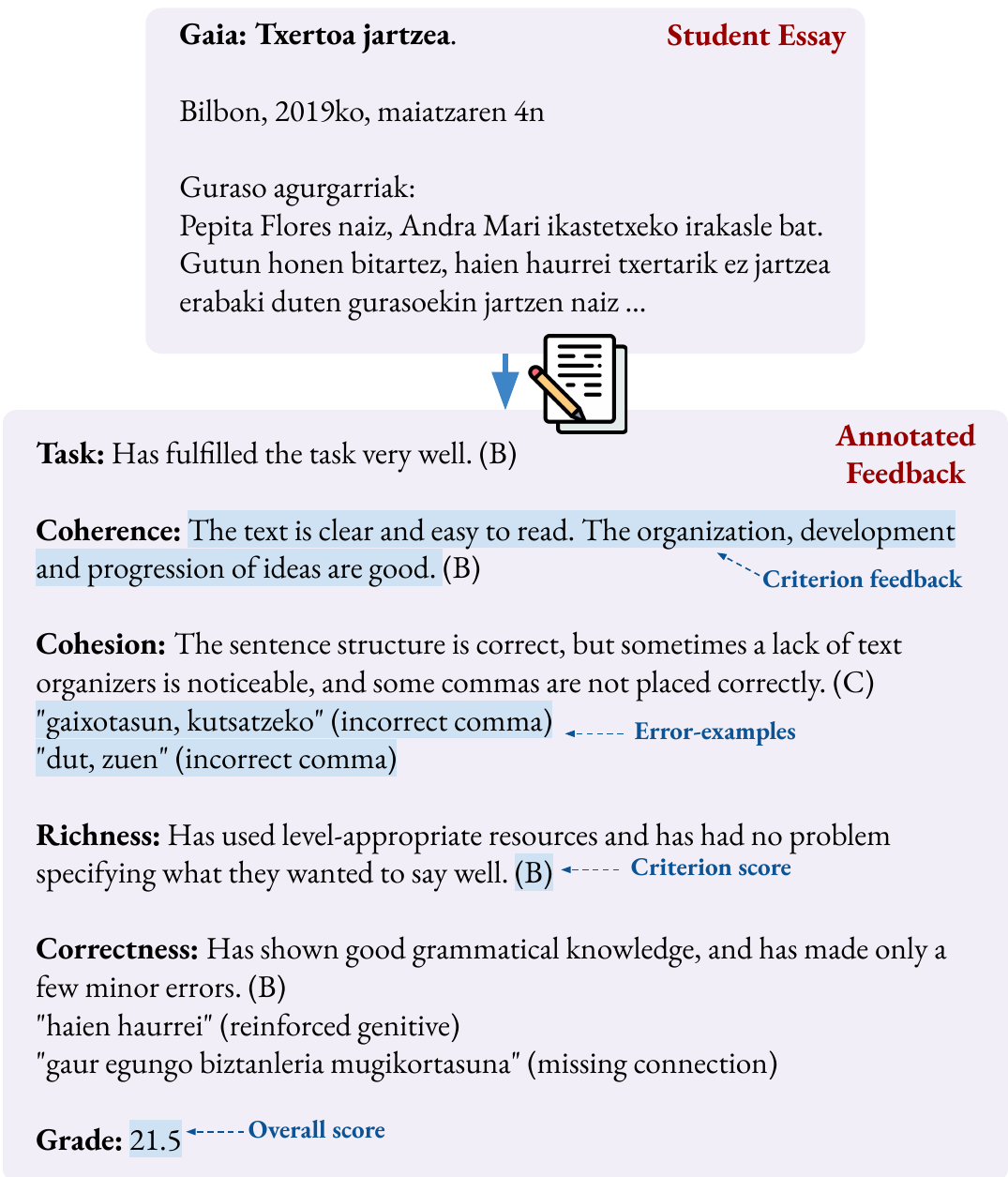}
    \caption{Excerpt from the Basque C1 dataset. Each essay includes multiple annotations, where each criterion (task alignment, coherence, lexical richness, and correctness) is aligned with natural-language feedback (\textbf{criterion feedback}), a score from A to D (\textbf{criterion score}), and learner error (error-example). Finally, an \textbf{overall score} of the essay is provided.}
    \label{fig:dataset}
\end{figure}

To mitigate this limitation, recent work has shifted towards \textbf{automated feedback generation}, aiming to complement the scoring with formative, interpretable, and pedagogically actionable insights that facilitate learner improvement.  Generating feedback on the essay level remains relatively unexplored beyond the grammatical errors at the sentence level ~\cite{nagata2019toward,song2023gee}.

Despite recent progress, most existing feedback generation methods remain dependent on closed-source proprietary models, which limit adaptability to task-specific assessment criteria and require extensive prompt engineering~\cite{stahl2024exploringllmpromptingstrategies}. The opacity of these systems hinders transparency, interpretability, and reproducibility. These limitations highlight the need for open frameworks that can produce accurate and pedagogically grounded feedback that aligns with established language proficiency scales and educational objectives. 

To overcome this limitation, we investigate strategies for fine-tuning Latxa \cite{etxaniz-etal-2024-latxa, sainz2025instructinglargelanguagemodels}, the Basque open-source generative LLM, for both AES and automated feedback generation in Basque. To this end, we have compiled a dataset of 3,200 Basque essays annotated with CEFR Level C1 scores, along with manually curated comments and explanations for incorrectly written sentences ~\footnote{Dataset and models are available in HugginFace: \url{https://huggingface.co/collections/EkhiAzur/habe-hitz-c1}}. The dataset provides detailed evaluations for each scoring criterion, including correctness, richness, coherence, cohesion, and task alignment, together with comments in natural language, and an aggregate overall essay score. This rich annotation enables the development of models capable of producing both reliable scores and interpretable, criterion-specific feedback. Figure~\ref{fig:dataset} illustrates an excerpt from the dataset, showing an essay as input, its corresponding criterion-wise feedback in natural language, associated scores, and the erroneous sentences produced by the learner. We argue that explicitly identifying learner errors alongside specific feedback offers substantial pedagogical value, fostering deeper awareness and more targeted language development. Although every criterion was fully annotated, in this paper we focused on correctness in feedback generation as first step, leaving the other criteria for future work.

Evaluation of feedback generation is non-trivial as it requires being assessed against well-defined rubrics. To address this, we designed a methodology for evaluating the explanations generated by the models. In our experiments, the quality of the feedback generation (i.e., the criterion feedback shown in Figure~\ref{fig:dataset}) was assessed automatically, while the models’ ability to identify learner errors (i.e., error examples in the figure) was evaluated by expert annotators. As a result of our experiments on AES and automated feedback generation, we find that:

\paragraph{1) Encoder-based classifiers remain competitive for AES.}
Our results show that a RoBERTa-EusCrawl–based~\cite{artetxe2022does} classifier outperforms the Latxa-8B model by a substantial margin across all evaluation criteria. To adapt Latxa to the task, we employed a 3-shot in-context learning setup. Despite the flexibility of large generative models, fine-tuned encoder architectures demonstrate great reliability and consistency in Automatic Essay Scoring.
\paragraph{2) Supervised fine-tuning (SFT) on Latxa improves correctness scoring.}
We compare the performance of a supervised fine-tuned Latxa model with RoBERTa and state-of-the-art (SoTA) systems, including GPT-5 and Sonnet 4.5. The SFT Latxa model achieves superior results in correctness scoring, demonstrating that training on the Basque C1 dataset effectively enhances model performance and domain adaptation.
\paragraph{3) Feedback and error-example generation enhance correctness scoring.}
We explore different strategies for generating scores, feedback, and error examples, and find that conditioning feedback and error generation on the predicted scores leads to improved performance in correctness scoring.
\paragraph{4) SFT Latxa demonstrates strong alignment between feedback and scores.}
The SFT Latxa model exhibits higher alignment (consistency) when providing correctness feedback compared to other models. In contrast, GPT-5 and Sonnet 4.5 show poor alignment between their feedback and the corresponding correctness scores, highlighting the advantage of SFT for reliable, criterion-consistent feedback.
\paragraph{5) Training on the dataset improves coverage of error-example types.} GPT-5 achieves similar overall scores on error identification. However, it primarily identifies spelling and lexical errors. In contrast, the SFT Latxa model detects errors across a broader range of categories. We conclude that the Basque C1 dataset enhances the model’s capability to extract diverse and pedagogically relevant error examples.

\begin{figure*}[!t]
    \centering
    \begin{subfigure}[b]{0.45\textwidth}
        \centering
        \includegraphics[width=\textwidth]{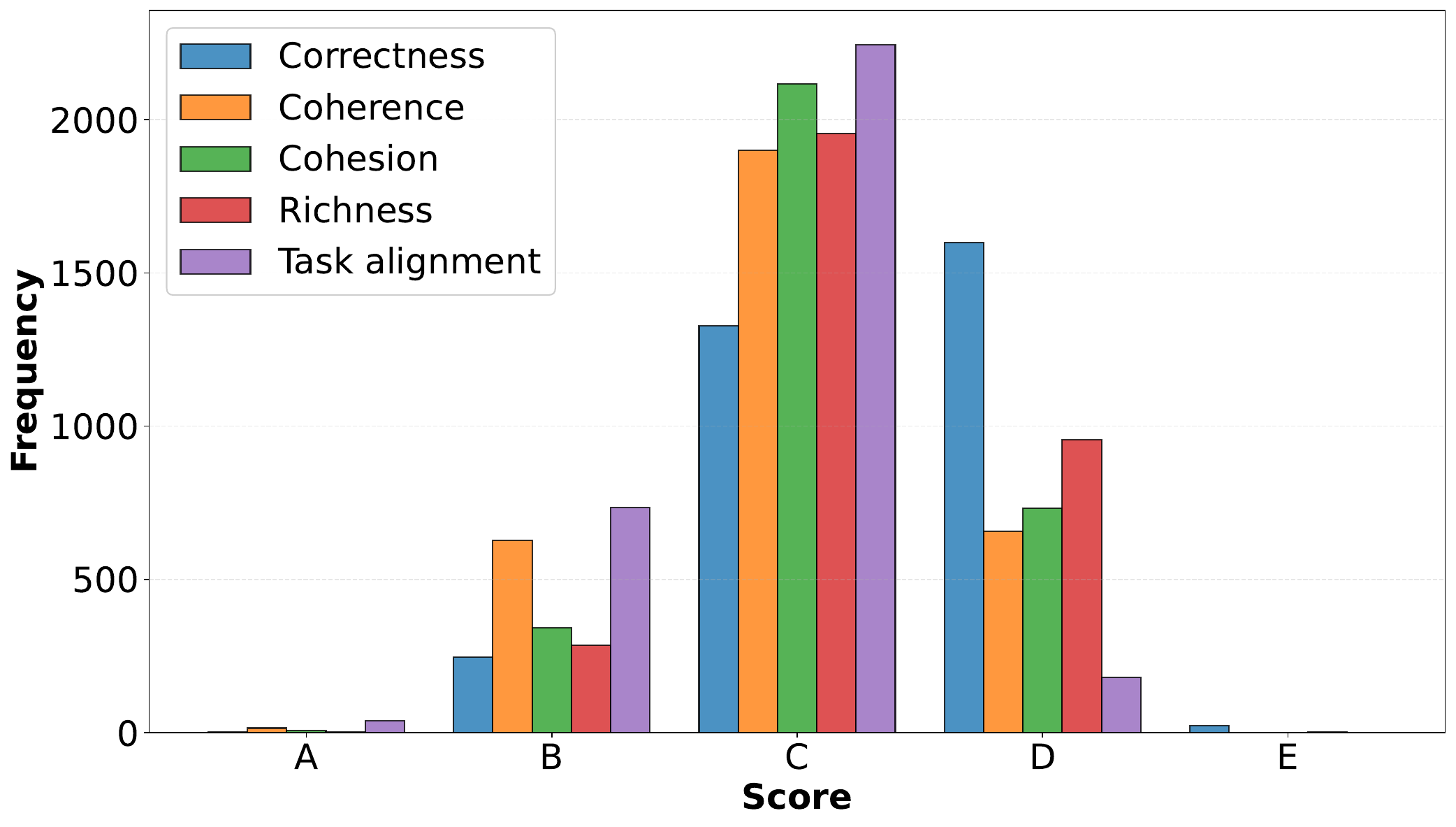}
        \caption{Score distribution across the evaluation criteria}
        \label{fig:label_distribution}
    \end{subfigure}
    \hfill
    \begin{subfigure}[b]{0.45\textwidth}
        \centering
        \includegraphics[width=\textwidth]{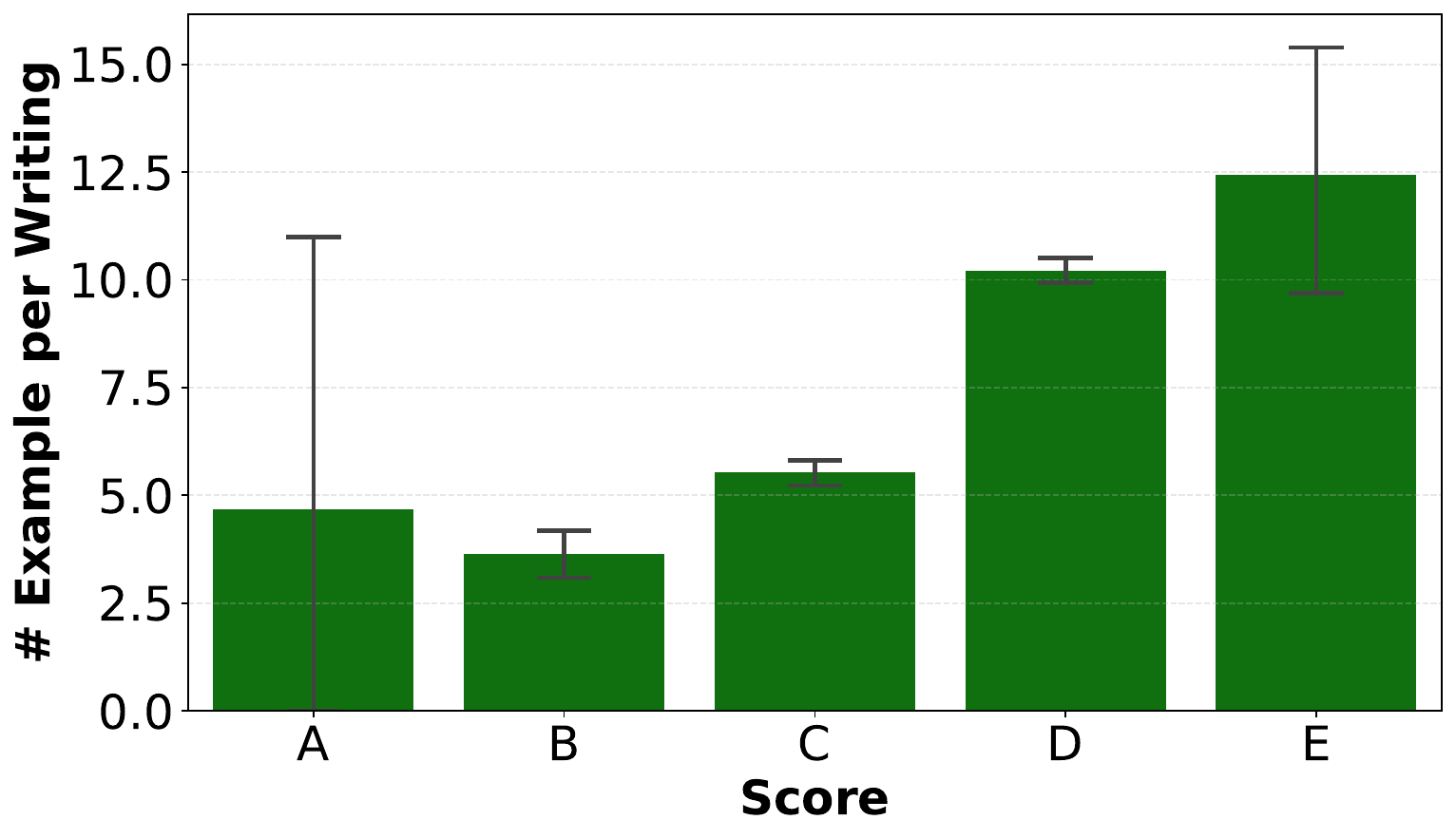}
        \caption{Annotated error-examples for correctness scores}
        \label{fig:adib_kop_nota}
    \end{subfigure}
    \caption{Basic statistics of the Basque C1 dataset.}
    \label{fig:datase_stats}
\end{figure*}

\section{Related work}

\paragraph{Automatic Essay Scoring}

Automatic Essay Scoring (AES) is the task that aims to automatically assess a score in an essay. The score range is predefined for each dataset. Although several works have been done in AES, it is still a challenging task with limited availability of datasets. Most studies focus on English language and little work has been done for other languages, mainly due to the lack of datasets for non-English languages.

Automated Student Assessment Prize (ASAP) \citep{ASAP} and TOEFL11 \citep{blanchard2013toefl11} are standard evaluation datasets in AES. They contain L2 English essays. ASAP contains eight essay sets written by students aged 12-16 years old, each set has its own score scale, and the essays were manually evaluated by two raters. TOEFL11 was created by collecting 12,000 english writings using 8 different prompts and were manually evaluated on a three-point scale.

Various strategies have been proposed to predict essay scores. Early approaches relied on linguistic features \citep{yannakoudakis2011new}, later replaced by neural systems \citep{dong2017attention}. During the recent years, transformers based models currently achieve SoTA performance, obtaining high accuracy \cite{yang-etal-2020-enhancing}.

Recently, research has been conducted investigating the effectiveness of generative LLMs in AES. \citet{mizumoto2023exploring} prompted GPT-3 to predict an overall score and scores for each predefined linguistic feature using TOEFL11 dataset, showing promising results. \citet{lee-etal-2024-unleashing} improved this technique by adding multi-turn prompts to extract scores for different linguistic features and lately adding them to get the final score. Although these studies achieve promising results, they rely on closed-source models and complex prompting strategies to obtain good results.

Despite the recent success of generative LLMs, encoder based models remain very capable for AES, achieving comparable or even superior results to very large LLMs \cite{xiao2025human, li-ng-2024-automated}.

\paragraph{Automatic Review Generation}

Recently, due to the strong performance of LLMs in NLP tasks, researchers have focused on using these models to generate reviews of different texts. \citet{zhou2024llm} explored the capability of GPT-3.5 and GPT-4 to assign scores and generate reviews of research papers. However, the results showed that further research is needed in order to enhance these models' performance.

Recent studies have focused on training LLMs to follow predefined guidelines to evaluate writings. \citet{kim2023prometheus} automatically created a dataset containing guidelines, writings and scores to fine-tune a model capable of generating reviews and predicting scores according to the guidelines. They achieved a high correlation with GPT-4 and human evaluators.

In the educational context, \citet{stahl2024exploringllmpromptingstrategies} explored different prompting strategies not only to assign scores but also to generate helpful reviews using Mistral \cite{jiang2023mistral7b} and Llama2 \cite{touvron2023llama2openfoundation} models. They experiment with role-based prompts and different output orders for scores and review  to improve performance. They found that addressing AES and review generation jointly is beneficial. Despite obtaining promising results they rely on zero-  or few-shot prompting without performing any fine-tuning.

\citet{nkoyoadvances} concluded that hybrid approaches, where the LLMs generated detailed feedback and predicted preliminary scores while humans handle more complex judgments are more effective rather than attempting to fully replace human evaluators.

Although some research has been done on review generation, there is no a standard evaluation framework and each study adopts its own evaluation methodology.

\paragraph{Basque Approaches in Education}

NLP in education suffers from a lack of annotated data \cite{li-ng-2024-automated}. This phenomenon is even worse in the case of Basque where, to the best of our knowledge, there is no a publicly available AES or review generation dataset.

Despite this data scarcity, some research has been done related to text correctness and CEFR (Common European Framework of Reference for Languages) level classification.

Inspired by BLiMP \cite{warstadt2020blimp}, \citet{urbizu2024well} created a similar dataset for Basque to evaluate the grammatical error detection capabilities of the models. They demonstrated that when presented with correct-incorrect sentence pairs, models are generally able to identify the erroneous sentence.

\citet{arriola-etal-2023-towards} used a private dataset to build a system for classifying writings according to CEFR levels, using linguistic features and Support Vector Machines (SVMs) \cite{cortes1995support}.

To the best of our knowledge, ours is the first publicly available Basque dataset for AES and feedback generation with error explanations, providing a crucial foundation for future research in educational applications.

\section{The Basque C1 Dataset}
\label{sec:dataset}

The Basque C1 dataset comprises essays written for HABE’s (Basque Government Department for Language Certification) C1 examinations. For each essay, HABE provides scores for the criteria used in the official evaluation. The evaluation criteria are as follows:

\begin{description}
    \item[Correctness:] Correctness is the accurate use of morphosyntax, vocabulary, spelling, and punctuation in writing, without systematic errors.
    \item[Richness:] Richness is the ability to write with a rich use of linguistic resources (vocabulary, structures, and ideas) demonstrating strong language knowledge.
    \item[Coherence:] Coherence is the clear and well-structured organization of a text, where main ideas and details are properly distinguished.
    \item[Cohesion:] Cohesion is the effective use of sentence, paragraph, and text-level links (discourse markers, cohesive devices, and punctuation).
    \item[Task Alignment:] Task Alignment is the ability to choose and use the appropriate register, vocabulary, and expressions according to the topic, context, audience, and purpose.
\end{description}

The HEOC document \cite{HEOC}, the adult Basque learner curriculum\footnote{\url{https://www.habe.euskadi.eus/contenidos/informacion/curriculuma/eu_9716/adjuntos/HEOC-2021-DIGITALA.pdf}}, provides additional details and explanations about these criteria.

Our dataset consists of 3,200 handwritten essays, which were automatically digitized by HABE using an OCR system with a mean Character Error Rate (CER) of 3.01\%. Manual review was performed to ensure text readability. Each essay was annotated by HABE’s professional evaluators according to five evaluation criteria, with each criterion scored on a scale from A to E, where A represents the highest performance. The final essay score was calculated as a weighted sum of these criterion-specific scores. This initial version of the dataset is suitable for AES tasks; however, it does not include explanations or suggestions for improving the essays.

\paragraph{Feedback and error-example annotation}
We expanded our original dataset by adding manually annotated explanations of the errors made by the examinee, providing not only a score but also detailed feedback about the performance of the student. These explanations include a general feedback and extractions of incorrectly written excerpts (typically sentences) from the essays and their error categorization for each criterion (we will refer to them as \textit{error-examples}). Note that the evaluator reports only some of the most relevant or significant errors made by the examinee, rather than every error. The evaluator decided the number of error examples to report, with no restrictions on the minimum or maximum number. Figure \ref{fig:dataset} shows an example of annotated feedback, error-examples, and scores for each criterion. The lower the score, the more error-examples are annotated (Figure~\ref{fig:adib_kop_nota} shows the number of annotated error-examples per correctness score). Note that the high variability in the number of examples extracted for the \textit{A}-score is due to the small number of \textit{A}-scored essays in the dataset.

We excluded the \textbf{Task Alignment} criterion, as it requires the essay prompt (which was not available in our dataset) to evaluate it properly and our analysis does not aim to evaluate discourse-level structures (e.g., letters, opinion essays).

Given the essay and the scores for each criterion, the feedback and error-examples were manually written by professional C1 evaluators from Hitzez Euskaltegia \footnote{\url{https://hitzez.eus/}} and Urrats Euskaltegia \footnote{\url{https://www.urrats.eus/}}. Generic annotation guidelines were defined to ensure consistency with the structure shown in Figure \ref{fig:dataset}, but we gave them some flexibility to have more variability in the annotations.

\paragraph{General overview}

The final version of the dataset contains 3,200 essays with scores and feedback for each criterion. %
The essays have an average length of 334.29 words. Figure~\ref{fig:label_distribution} shows the score distributions for the criteria, which are generally centered on label C, except for \textit{Correctness}, which is centered on label D. %

Table~\ref{tab:explanations_description} shows the average length of the feedback in words along with the number of annotated error examples per criterion. Among all the criteria, \textit{correctness} shows the highest number of annotated error examples. This is expected as it is the most objective criterion. \textit{Richness} and \textit{Coherence} are the most abstract criteria; therefore, it is more difficult to identify specific errors in the text, resulting in fewer extracted examples. The number of examples in \textit{Correctness} is correlated with the writing score (See Figure \ref{fig:adib_kop_nota}). Writings with higher scores tend to contain fewer errors, thus the present fewer error examples.

\begin{table}[!t]
\begin{tabular}{l|cc}
\toprule
\textbf{Criterion} & \textbf{Feedback length} & \textbf{\# Example} \\ \midrule
Correctness & 15.63 $\pm$ 8.82 & 7.78 $\pm$ 6.19 \\
Richness & 23.31 $\pm$ 19.20 & 0.62 $\pm$ 1.39 \\
Coherence & 26.64 $\pm$ 20.92 & 0.13 $\pm$ 0.49 \\
Cohesion & 21.32 $\pm$ 15.47 & 2.94 $\pm$ 3.41 \\
\bottomrule

\end{tabular}
\caption{Feedback length and number of error examples per criteria.}
\label{tab:explanations_description}
\end{table}

The dataset was divided into training, validation, and test sets that contained 2600, 300, and 300 instances, respectively.

\section{Experimental setting} 
\label{sec:experimental_setting}

We conducted two types of experiments. The first focuses on AES, aiming to predict the score for each criterion, while the second jointly generates the explanations and the corresponding scores.

\paragraph{AES setting} We fine-tuned RoBERTa-EusCrawl large \cite{artetxe2022does} and evaluated Latxa 8B Instruct \cite{sainz2025instructinglargelanguagemodels} using EleutherAI's lm-evaluation-harness \cite{eval-harness} for each evaluation criterion. The models were fine-tuned with a learning rate of $5e-5$ using a cosine scheduler, weight decay of $5e-3$, and a batch size of 32 during 10 epochs. No hyperparameter optimization was performed. To evaluate Latxa we used a 3-shot evaluation.

\paragraph{Feedback Generation setting} For the experiments, we focused only on \textit{Correctness} due to the high computational cost of running all experiments. We focus on \textit{Correctness}, as it is the most objective and contains more error examples compared to the rest of the criteria. We plan to explore additional criteria in the future.

For model training, we conducted full fine-tuning using SFT, computing the loss exclusively on the model outputs. The training employed a learning rate of $5\times10^{-6}$ with a cosine learning rate scheduler, a weight decay of 0.1, and a batch size of 64 over 10 epochs. DeepSpeed ZeRO \cite{rajbhandari2020zeromemoryoptimizationstraining} optimization was applied, using Stage 2 for the 8B models and Stage 3 for the 70B model.

We explored the effect of different orderings of score (S), feedback (F), and error-examples (E) as the predicted output by fine-tuning separate Latxa 8B Instruct models. We discarded configurations that did not include the score in the output, as we used the scores to automatically select the best configuration\footnote{We evaluated seven different orderings in total: S, SF, FS, SFE, SEF, EFS, and ESF.}. Best configuration was selected for training the Latxa 70B Instruct model~\cite{sainz2025instructinglargelanguagemodels}. %

As baselines, we evaluated Latxa 70B Instruct without fine-tuning, GPT-5\footnote{gpt-5-2025-08-07}, and Claude Sonnet 4.5\footnote{claude-sonnet-4-5-20250929} using their best-performing configurations for comparison with our top model. Due to the extensive context length required by the prompt, the evaluation was limited to a 1-shot setting.

\section{Evaluation methodology} \label{sec:evaluation_methodology}

Our models generate up to three types of outputs: score, feedback, and error-examples. We evaluated the score and feedback output using automatic metrics, while the error-examples were assessed by human annotators.

\subsection{AES Evaluation Metrics} \label{subsec:AES_metrics}

We have used Quadratic Weighted Kappa (QWK) \cite{cohen1968weighted} as the main metric for scoring prediction as it is the standard metric used in AES \cite{li-ng-2024-automated}. We have also calculated Weighted-F1 as it is especially interesting as our dataset is unbalanced in the extreme labels (see Table \ref{fig:label_distribution}). We also calculated the Pearson correlation to measure the correlation between our systems and professional C1 evaluators.

\subsection{Feedback Generation Evaluation} \label{subsec:review_generation_evaluation}

\paragraph{Automated Feedback Evaluation} %
The evaluation was conducted by mapping the generated feedback to predicted scores using the RoBERTa EusCrawl Large model. We assume that if the predicted scores show a strong correlation with human ratings, the generated feedback can be regarded as high-quality. We computed the QWK between the predicted and actual scores, defining this metric as \textbf{consistency}. The model was fine-tuned using the same hyperparameters described in Section~\ref{sec:experimental_setting}. It achieved a QWK of 0.776 and a Weighted-F1 score of 89.0, demonstrating a strong ability to accurately map textual feedback to corresponding scores.

\paragraph{Manual Evaluation} %

We propose a framework to assess models’ ability to identify and categorize representative examples of errors made by examinees. Specifically, the model must reproduce each error example accurately, without introducing any modifications. To enable a more detailed analysis of model performance, we classify errors into seven distinct categories: spelling, incorrect declensions, auxiliary verb errors, morphological errors, syntactic errors, inappropriate vocabulary usage, and punctuation errors.

A total of eight native Basque speakers with backgrounds in linguistics and philology participated as evaluators, some of whom currently serve or have previously served as professional evaluators at HABE. The evaluators were asked to answers the following for each example (See full guidelines in \ref{apx:full_guidelines}: 

\begin{description}
    \item[Q1:] Does the sentence exist in the text? or is it a hallucination?
    \item[Q2:] Is the extracted sentence wrongly written?
    \item[Q3:] Classify the error type given by the model into our categories.
    \item[Q4:] Does the model detect the error correctly?
\end{description}

Based on these questions, we designed 3 metrics to evaluate the systems' capabilities in extracting and categorized erroneous sentences:

\begin{description}
    \item[Fidelity Rate (FR)] measures the percentage of extracted sentences that exist in the writing. Higher is better.
    \item[Extraction Accuracy (EA)] measures the percentage of extracted sentences that contain errors made by learners. Higher is better.
    \item[Categorization Accuracy (CA)] measures the accuracy of error categorization. This metric is computed using Micro-F1. Higher is better.
\end{description}

Due to the high cost of human evaluation, we limited the manual evaluation to our best performing Latxa 70B model and the 3 %
baseline models.

\section{AES Results}

\begin{table}[!t]
\begin{small}
\centering
\begin{tabular}{l|l|rrr}
\toprule
\textbf{Criterion} & \textbf{Model} & \textbf{QWK} & \textbf{W-F1} & \textbf{Pearson} \\
\midrule
\multirow{2}{*}{Task align.} & R-Eusc & \textbf{17.85}& \textbf{65.7}\textbf{8}& \textbf{19.08}\\ 
 & Latxa 8B  & 3.35& 40.25& 4.49\\ 
 \midrule
\multirow{2}{*}{Correctness} & R-Eusc & \textbf{43.82}& \textbf{62.51}& \textbf{44.94}\\
 & Latxa 8B  & 5.16& 27.16& 9.00\\ \midrule
\multirow{2}{*}{Richness} & R-Eusc & \textbf{18.31}& \textbf{50.05}& \textbf{21.97}\\
 & Latxa 8B  & 2.19& 27.91& 3.76\\ \midrule
\multirow{2}{*}{Coherence} & R-Eusc & \textbf{9.27}& \textbf{54.46}& \textbf{10.81}\\
 & Latxa 8B  & 6.11& 47.72& 7.94\\ \midrule
\multirow{2}{*}{Cohesion} & R-Eusc & \textbf{19.71}& \textbf{65.94}& \textbf{21.69}\\
 & Latxa 8B  & 4.72& 51.46& 6.46\\ 
\bottomrule
\end{tabular}
\caption{Performance of RoBERTa EusCrawl (R-Eusc in the table) and Latxa 8B in AES for each criterion. W-F1 stands for weighted F1 score. }
\label{tab:all_criteria_AES_results}
\end{small}
\end{table}

\paragraph{Overall AES results}
A comparison of the fine-tuned RoBERTa-EusCrawl models against the 3-shot Latxa 8B baseline across all evaluation criteria shows that Latxa achieves consistently lower performance (see Table \ref{tab:all_criteria_AES_results}). Latxa 8B obtains near zero QWK values, indicating no capability to assess scores using 3-shot evaluation. The highest QWK value is obtained in \textit{Correctness}, showing that this particular criterion is relatively easier than the other criteria. Furthermore, the discrepancy between high Weigthed-F1 scores compared to lower QWK values across the other criteria is probably due to the imbalanced scores presented in the dataset (see Figure \ref{fig:label_distribution}).

\paragraph{Basque SoTA in Correctness} %
Table~\ref{tab:SOTA_AES_results} presents the evaluation of Basque SoTA models under a 1-shot prompting setup, and compares with our SFT Latxa 70B, which was evaluated in a zero-shot setting. Our fine-tuned model achieves the highest performance, with a QWK of 57.23, surpassing the best encoder-based models by 13 points (43.82, Table~\ref{tab:all_criteria_AES_results}). Among the non–fine-tuned systems, the 1-shot Latxa 70B marginally outperforms private models by 3–4 QWK points. These results indicate that applying SFT on our dataset substantially enhances the model’s assessment capabilities, yielding a 35-point improvement over the base Latxa 70B. Moreover, the significantly stronger performance of SoTA models compared to the 3-shot Latxa 8B confirms that larger generative models exhibit superior ability in score assessment tasks.

\begin{table}[!t]
\centering
\begin{tabular}{l|rrr}
\toprule
  \textbf{Model} & \textbf{QWK} & \textbf{W-F1} & \textbf{Pearson} \\ \midrule
  Latxa-it 70B & 21.37& 51.27& 27.23\\
   Claude Sonnet 4.5 & 18.93& 45.02& 22.60\\
   GPT5 & 17.59& 48.00& 19.94\\ \midrule
   SFT Latxa 70B*  & \textbf{57.23}& \textbf{69.97}& \textbf{57.82}\\ 
\bottomrule
\end{tabular}
\caption{Performance of basque SOTA generative models in assessing Correctness scores. We used SFE output ordering and 1-shot prompting. *SFT Latxa 70B was prompted using zero-shot. W-F1 stands for Weighted F1 score.}
\label{tab:SOTA_AES_results}
\end{table}

While the score assessment results of non fine-tuned generative models are lower than the fine-tuned encoder models, the use of generative models offers the advantage to generate feedback and give explanations of identified errors. These explanations benefit the user of these systems, serving as a strong incentive to adapt these classification tasks to generative LLMs.

\paragraph{Analysis of Output Configuration}
Table~\ref{tab:sft_AES_results} shows the influence of the output ordering in AES (details in Section~\ref{sec:experimental_setting}). The results indicates that output configuration significantly affects model performance, with the SFE ordering being the best configuration. 

The configuration that prioritizes initial score assessment is consistently optimal, whereas performance is notably lower when the score prediction occurs after the generation of feedback or error examples\footnote{We did not evaluate the FES and FSE configurations as the configurations that initially predict feedback generation obtained lower  performance.}. We hypothesize that predicting first score (S) enable the model to adequate the following generation of feedback and error examples. 
These findings indicate that the fine-tuned 8B models are capable of surpassing the performance of larger models, both private and open-source, if the 8B models are fine-tuned specifically for this task. Consistent with our previous observations on 70B model, SFT substantially enhances model performance, increases from 5.16 (Table~\ref{tab:all_criteria_AES_results}) to 36.60 (Table~\ref{tab:sft_AES_results}) for the Latxa 8B model in QWK using identical output configuration (S).

\begin{table}[!t]
\centering
\begin{tabular}{l|l|rrr}
\toprule
  \textbf{Size} &\textbf{Output} & \textbf{QWK} & \textbf{W-F1} & \textbf{Pearson} \\ \midrule
  \multirow{8}{*}{8B}&S & 36.60& 58.35& 37.50\\
   &SF & 30.64& 55.46& 31.43\\
   &FS & 24.14& 53.54& 24.31\\
   &ESF & 28.42& 54.72& 29.63\\
   &EFS & 16.70& 51.68& 17.17\\
   &SFE & \textbf{39.52}& \textbf{62.39}& \textbf{41.67}\\
   &SEF & 38.07& 59.92& 39.07\\ \midrule
  70B  &SFE & \textbf{57.23}& \textbf{69.97}& \textbf{57.82}\\
  \bottomrule
\end{tabular}
\caption{Performance of SFT Latxa 8B and SFT Latxa 70B in Correctness score assessing. W-F1 stands for Weighted F1 score.}
\label{tab:sft_AES_results}
\end{table}

Scaling the optimal SFE configuration to the Latxa 70B model yields a substantial 17.7-point gain in QWK, enabling it to surpass encoder models in \textit{Correctness} prediction while also generating error explanations and categorizations. In contrast, fine-tuned Latxa 8B models for \textit{Correctness} assessment perform worse than fine-tuned RoBERTa-EusCrawl encoders. This trend, where fine-tuned generative models underperform compared to encoder models, was also noted by \citet{xiao2025human}, who observed similar results with their fine-tuned GPT-3.5 and Llama-3 models.

\section{Feedback Generation Results}

The error explanations that generate our models are composed by two elements: a short feedback and error-examples. 

\subsection{Feedback Consistency Evaluation}

We evaluate the short feedback using the Consistency metric, as detailed in Section \ref{sec:evaluation_methodology}. All fine-tuned Latxa models, both 8B or 70B, achieved exceptionally high Consistency values, with the minimum score being 94.1 (see Table \ref{tab:Consistency metric}). Although the SFE configuration gets slightly lower Consistency results compared to other output sequences, the difference is marginal suggesting that SFE maintains high consistency in both score and feedback generation. These results may suggest that improving the score assessing capabilities of SFT models will consequentially enhance the quality of these short feedback.

In contrast, the results for SoTA models are substantially lower, with GPT5 obtaining the lowest scores with 44.1. Non fine-tuned Latxa 70B obtains higher \textit{Consistency} value than other non-fine-tuned closed-source models, achieving similar scores of fine-tuned models with a margin of 8 points.

\begin{table}[!t]
    \centering
    \begin{tabular}{l|l|r}
    \toprule
         \textbf{Model} &\textbf{Output} & \textbf{Consistency} \\ 
         \midrule
          \multirow{7}{*}{SFT Latxa 8B} &SF & 94.07\\
          &FS & 94.73\\
          &ESF & 96.87\\
          &EFS & 96.12\\
          &SFE & 94.28\\
          &SEF & \textbf{97.33}\\ \midrule
          SFT Latxa 70B &SFE& \textbf{96.84}\\ \midrule
          Latxa 70B&SFE& 86.20\\
          GPT5&SFE& 44.07\\
          Sonnet 4.5&SFE& 78.46\\
        \bottomrule
    \end{tabular}
    \caption{Results of Consistency of Correctness feedback.}
    \label{tab:Consistency metric}
\end{table}

\subsection{Manual Evaluation}

Manual evaluation of error examples followed the annotation methodology in Section~\ref{subsec:review_generation_evaluation}. Inter-annotator agreement was very high for the first question (95.74\%), and progressively lower for the following ones (71.28\%, 68.33\%, and 68.09\%), reflecting the increasing complexity of the annotations.

The results of the manual evaluation are presented in Table \ref{tab:manual_results}. All the models achieve high values on Fidelity Rate (FR), showing a strong capability to extract sentences without hallucination, with GPT-5 obtaining the highest score of 100\%. Regarding Extraction Accuracy (EA) and Categorization Accuracy (CA), private models outperformed both Latxa variants, with GPT-5 again achieving the highest accuracy scores in both metrics. 

\begin{table}[h]
\centering
    \begin{tabular}{l|rrr}
        \toprule
        \textbf{Model} & \textbf{FR} & \textbf{EA} & \textbf{CA} \\ \midrule
        SFT Latxa 70B & 98.08 & 66.19 & 71.63 \\ \midrule
        Latxa & 97.64 & 61.08 & 51.96 \\
        GPT5 & \textbf{100.0} & \textbf{81.88} & \textbf{82.47} \\
        Sonnet 4.5 & 98.79 & 72.54 & 78.61 \\ \bottomrule
    \end{tabular}
    \caption{Results of models in manual evaluation. Fidelity Rate (FR), Extraction Accuracy (EA) and Categorization Accuracy (CA) are defined in Section \ref{subsec:review_generation_evaluation}}
    \label{tab:manual_results}
\end{table}

Fine-tuning in our dataset significantly boosted the performance of Latxa 70B across every manual evaluation metric, decreasing the performance gap between open-source and closed-source models.

\begin{figure*}[!t]
    \centering    
    \includegraphics[scale=0.42]{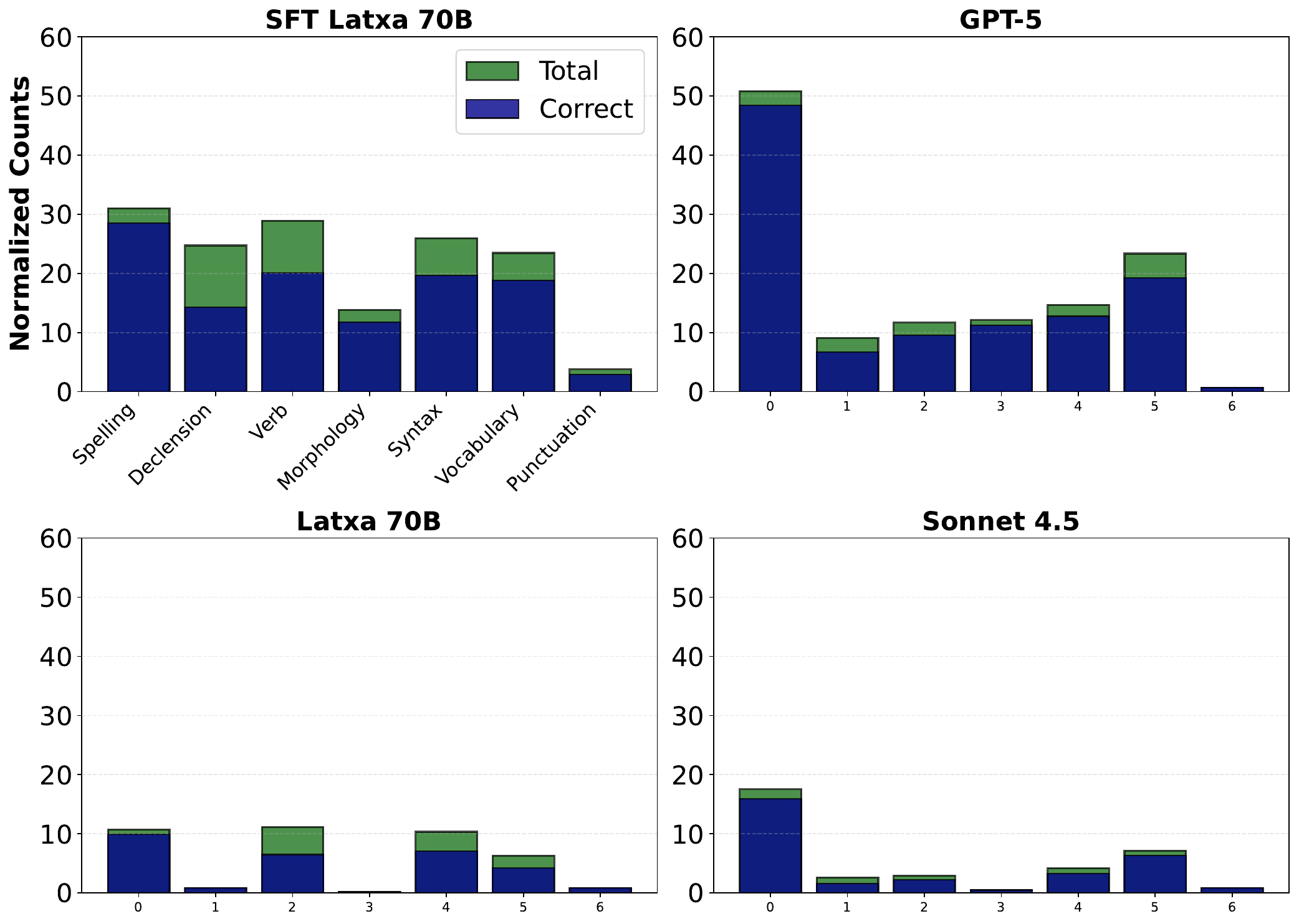}
    \caption{Category percentage and accuracy of models}
    \label{fig:manual_category}
\end{figure*}

Figure \ref{fig:manual_category} shows the error categorization distribution (green) of each model and the accuracy per category (blue), which is normalized by the total number of 
annotated essays per model. The category distribution of the fine-tuned Latxa 70B is more balanced compared to other models, showing competitive capabilities against GPT-5. GPT-5 model has a strong tendency to extract Spelling and Vocabulary category related errors, probably due to OCR errors present in the essays. OCR related Spelling and Vocabulary errors are easier to detect and classify, making it easier to achieve higher accuracy in these categories. However, our dataset contains few error-examples related to OCR errors as they are less pedagogically relevant than other error categories.

We observe that non SFT Latxa 70B and Sonnet 4.5 extract significantly fewer sentences than GPT-5 and fine-tuned Latxa. Although both closed-source models differ in category extraction, they show only a small gap between the distribution (green) and accuracy (blue) bars, indicating that, especially Sonnet 4.5, they take few risks in extraction. This cautious behavior may explain the lower accuracy of SFT Latxa, which takes greater risks to cover all categories. Overall, SFT Latxa shows substantial improvement, achieving higher scores across all manual evaluation metrics and covering a wider range of categories compared to non SFT Latxa, which extracts fewer sentences with lower accuracy.

\section{Conclusions}

This paper introduced the first publicly available, richly annotated dataset for Basque AES and feedback generation at the CEFR C1 level. Our experiments demonstrate that while fine-tuned encoder models like RoBERTa-EusCrawl remain a strong baseline for criterion-based AES, training generative models using SFT on our new dataset yields significant performance gains. Specifically, the SFT Latxa 70B model surpassed both specialized encoder models and state-of-the-art proprietary models like GPT-5 and Claude Sonnet 4.5 in score Correctness criterion.

Furthermore, our analysis of generated explanations revealed that SFT models have high consistency between generated feedback and assessed score. The fine-tuned Latxa model also proved superior in identifying a more balanced and pedagogically relevant range of error types, whereas closed models disproportionately focused on surface-level spelling and vocabulary errors, possibly due to OCR artifacts in the essays. As future work, we plan to expand our experimental analysis of feedback generation to all evaluation criteria beyond Correctness. We also intend to further investigate training techniques and evaluation methodologies that incorporate the pedagogical significance and recall of the error-examples, moving beyond simple accuracy to measure true educational value.

\section*{Limitations}

The current study primarily focused on evaluating the Correctness criterion; consequently, experiments covering the remaining scoring criteria remain unexplored. The "Task Alignment" criterion was also excluded, as the original essay prompts were unavailable. Furthermore, the use of OCR to digitize the handwritten essays (3.01\% CER) influenced error-example evaluation, as models sometimes identified OCR artifacts rather than authentic learner errors. The feedback generation was evaluated only in terms of its consistency with the predicted score, without assessing the quality of the feedback itself or comparing it to manually written feedback. Finally, manual evaluation do not analyze the pedagogical significance of error-examples nor compute the recall of the error-examples.

\section*{Ethical Consideration}

Regarding the personal data processed during this research, every essay and its metadata were previously anonymized by HABE during the data transfer process. Additionally, HABE warns writers to avoid including personal data in their essays, meaning no personal data is present in the dataset or was used during the research. HABE also holds the rights to use these essays and the anonymized metadata for research purposes, and every examinee must agree to these conditions.

\section*{Acknowledgements}

This work has been partially supported by the Basque Government (Research group funding IT1570-22 and IKER-GAITU project), the Spanish Ministry for Digital Transformation and Civil Service, the EU-funded NextGenerationEU Recovery, Transformation and Resilience Plan (ILENIA project, 2022/TL22/00215335), the European Union (EFA 104/01-LINGUATEC IA project) and the CLARIAH-ES and CLARIAH-EUS initiatives. Ekhi Azurmendi hold a PhD grant from the Basque Government (PRE\_2024\_1\_0035). The models were trained on the Leonardo supercomputer at CINECA under the EuroHPC Joint Undertaking, project EHPC-EXT-2024E01-042 and on Calendula supercomputer at SCAYLE, thanks to CLARIAH-ES.

\section*{Bibliographical References}\label{sec:reference}
\bibliographystyle{lrec2026-natbib}
\bibliography{lrec2026-example}

\appendix

\section{Manual Evaluation Guidelines} \label{apx:full_guidelines}

\subsection{Evaluation Questions (Per Example)}
For each extracted example, annotators must answer the following sequential questions:

\begin{enumerate}
    \item \textbf{Is the example present in the text? (Yes / No)}
    \begin{itemize}
        \item \textit{Note:} Some models may correct the OCR errors during the extraction. Ignore those corrections.
    \end{itemize}

    \item \textbf{Does the example contain an error? (Yes / No / Skip)}
    \begin{itemize}
        \item Answer \textbf{Yes} if the sentence is grammatically incorrect or if a significantly better grammatical structure exists for the C1 level.
        \item Answer \textbf{No} if the identified error is purely stylistic or not related to correctness.
    \end{itemize}

    \item \textbf{What is the identified error category?} 
    Identify the error category of the error-example: Orthography, Declension, Verb, Morphology, Syntax, Lexicon, or Punctuation.
    \begin{itemize}
        \item \textit{Note:} If the model outputs a subcategory (e.g., ``ergative''), map it to the correct main category (e.g., Morphology).
    \end{itemize}

    \item \textbf{Is the assigned category correct? (Yes / No)} \\
    Evaluate whether the model's predicted category matches the actual error in the text.
\end{enumerate}

\subsection{Error Categories}

Human annotators defined the error categories based on recurrent mistakes in Basque C1 exams. We distinguished declension from morphology due to its fundamental role in Basque grammar.

\begin{description}
    \item[Orthography:] Spelling mistakes (e.g., \textit{lehioa} $\rightarrow$ \textit{leihoa}).
    \item[Declension:] Incorrect case suffixes or postpositions (e.g., \textit{dentistara joan} $\rightarrow$ \textit{dentistarenera joan}).
    \item[Verb:] Agreement errors, incorrect tense, or wrong auxiliary choice (e.g., \textit{nik lagunari eman dut} $\rightarrow$ \textit{nik lagunari eman diot}).
    \item[Morphology:] Ergative case errors or number mismatch (e.g., \textit{nik etorri naiz} $\rightarrow$ \textit{ni etorri naiz}).
    \item[Syntax:] Word order issues, subordinate clause construction errors, or incorrect calques from Spanish.
    \item[Lexicon:] Incorrect vocabulary choices (e.g., \textit{urgentziak} $\rightarrow$ \textit{larrialdiak}).
    \item[Punctuation:] Misplaced commas, semicolons, or related punctuation errors that impact readability or syntax.
\end{description}

\section{Hyperparameters}

Table \ref{tab:apx_hyp_finetuning} describes the hyperparameters used to fine-tune the Latxa 8B and 70B models. We adapted the TRL \citep{vonwerra2020trl} framework to adapt the models using Flash-Attention-2 \citep{dao2023flashattention2} and Liger-Kernel \citep{hsu2025ligerkernel} optimized kernels to accelerate the training. We used DeepSpeed to optimally parallelize the training, using ZeRO2 to fine-tune Latxa 8B and ZeRO3 for the 70B model. 

We used vLLM v0.7 for generation with the default Latxa hyperparameters used by \cite{sainz2025instructinglargelanguagemodels}. For closed source models, we used a temperature value of 0.0 for reproducibility.

\begin{table}[]
    \centering
    \begin{tabular}{c|c}
        Hyperparam & Value \\ \hline
        Batch size & 64 \\
        Learning Rate & $5e-6$ \\
        Weight Decay & $0.1$ \\
        Epochs & $10$ \\
        Learning Rate Decay & Cosine \\
        Warmup ratio & $0.1$ \\
    \end{tabular}
    \caption{Hyperparameters for fine-tuning Latxa 8B and 70B}
    \label{tab:apx_hyp_finetuning}
\end{table}

\end{document}